%% file: main.tex
\documentclass[10pt,twocolumn,letterpaper]{article}

\usepackage{cvpr}              
\usepackage{array}
\usepackage[accsupp]{axessibility}
\input{preamble}

\definecolor{cvprblue}{rgb}{0.21,0.49,0.74}
\definecolor{mygreen}{HTML}{32963d}
\usepackage[pagebackref,breaklinks,colorlinks,citecolor=blue]{hyperref}

\def\confName{CVPR}
\def\confYear{2024}

\title{Second Edition FRCSyn Challenge at CVPR 2024:\\Face Recognition Challenge in the Era of Synthetic Data}

\author{Ivan DeAndres-Tame$^{1}$
\and 
Ruben Tolosana$^{1}$
\and
Pietro Melzi$^{1}$
\and 
Ruben Vera-Rodriguez$^{1}$
\and
Minchul Kim$^{2}$
\and
Christian Rathgeb$^{3}$
\and 
Xiaoming Liu$^{2}$
\and
Aythami Morales$^{1}$
\and
Julian Fierrez$^{1}$
\and 
Javier Ortega-Garcia$^{1}$
\and
Zhizhou Zhong$^{4}$
\and
Yuge Huang$^{5}$
\and
Yuxi Mi$^{4}$
\and
Shouhong Ding$^{5}$
\and
Shuigeng Zhou$^{4}$
\and
Shuai He$^{6}$
\and
Lingzhi Fu$^{6}$
\and
Heng Cong$^{6}$
\and
Rongyu Zhang$^{6}$
\and
Zhihong Xiao$^{6}$
\and
Evgeny Smirnov$^{7}$
\and
Anton Pimenov$^{7}$
\and
Aleksei Grigorev$^{7}$
\and
Denis Timoshenko$^{7}$
\and
Kaleb Mesfin Asfaw$^{8}$
\and
Cheng Yaw Low$^{9}$
\and
Hao Liu$^{10}$
\and
Chuyi Wang$^{10}$
\and
Qing Zuo$^{10}$
\and
Zhixiang He$^{10}$
\and
Hatef Otroshi Shahreza$^{11, 12}$
\and
Anjith George$^{11}$
\and
Alexander Unnervik$^{11, 12}$
\and
Parsa Rahimi$^{11, 12}$
\and
S\'{e}bastien Marcel$^{11, 13}$
\and
Pedro C. Neto$^{14, 15}$
\and
Marco Huber$^{16}$
\and
Jan Niklas Kolf$^{16}$
\and
Naser Damer$^{16}$
\and
Fadi Boutros$^{16}$
\and
Jaime S. Cardoso$^{14, 15}$
\and
Ana F. Sequeira$^{14, 15}$
\and
Andrea Atzori$^{17}$
\and
Gianni Fenu$^{17}$
\and
Mirko Marras$^{17}$
\and
Vitomir \v{S}truc$^{18}$
\and
Jiang Yu$^{19}$
\and
Zhangjie Li$^{19, 20}$
\and
Jichun Li$^{19}$
\and
Weisong Zhao$^{21}$
\and
Zhen Lei$^{22}$
\and
Xiangyu Zhu$^{22}$
\and
Xiao-Yu Zhang$^{21}$
\and
Bernardo Biesseck$^{23, 24}$
\and
Pedro Vidal$^{23}$
\and
Luiz Coelho$^{25}$
\and
Roger Granada$^{25}$
\and
David Menotti$^{23}$
\and
\normalsize{$^{1}$Universidad Autonoma de Madrid, Spain}
\normalsize{$^{2}$Michigan State University, USA}
\normalsize{$^{3}$Hochschule Darmstadt, Germany}\\
\normalsize{$^{4}$Fudan University, China}
\normalsize{$^{5}$Tencent Youtu Lab, China}
\normalsize{$^{6}$Interactive Entertainment Group of Netease Inc, China}\\
\normalsize{$^{7}$ID R\&D Inc., USA}
\normalsize{$^{8}$Korea Advanced Institute of Science \& Technology, Korea}
\normalsize{$^{9}$Institute for Basic Science, Korea}\\
\normalsize{$^{10}$China Telecom AI, China}
\normalsize{$^{11}$Idiap Research Institute, Switzerland}
\normalsize{$^{12}$EPFL, Switzerland}
\normalsize{$^{13}$Universit\'{e} de Lausanne, Switzerland}\\
\normalsize{$^{14}$INESC TEC, Portugal}
\normalsize{$^{15}$Universidade do Porto, Portugal}
\normalsize{$^{16}$Fraunhofer IGD, Germany}
\normalsize{$^{17}$University of Cagliari, Italy}\\
\normalsize{$^{18}$University of Ljubljana, Slovenia}
\normalsize{$^{19}$Samsung Electronics (China) R\&D Centre, China}\\
\normalsize{$^{20}$University of Science and Technology, China}
\normalsize{$^{21}$IIE, CAS, China}
\normalsize{$^{22}$MAIS, CASIA, China}\\
\normalsize{$^{23}$Federal University of Paraná, Brazil}
\normalsize{$^{24}$Federal Institute of Mato Grosso, Brazil}
\normalsize{$^{25}$unico - idTech, Brazil}
}
\begin{document}
\maketitle

\begin{abstract}

Synthetic data is gaining increasing relevance for training machine learning models. This is mainly motivated due to several factors such as the lack of real data and intra-class variability, time and errors produced in manual labeling, and in some cases privacy concerns, among others. This paper presents an overview of the 2$^\text{nd}$ edition of the Face Recognition Challenge in the Era of Synthetic Data (FRCSyn) organized at CVPR 2024. FRCSyn aims to investigate the use of synthetic data in face recognition to address current technological limitations, including data privacy concerns, demographic biases, generalization to novel scenarios, and performance constraints in challenging situations such as aging, pose variations, and occlusions. Unlike the 1$^\text{st}$ edition, in which synthetic data from DCFace and GANDiffFace methods was only allowed to train face recognition systems, in this 2$^\text{nd}$ edition we propose new sub-tasks that allow participants to explore novel face generative methods. The outcomes of the 2$^\text{nd}$ FRCSyn Challenge, along with the proposed experimental protocol and benchmarking contribute significantly to the application of synthetic data to face recognition.
\end{abstract}

\section{Introduction}\label{sec:intro}
\begin{figure*}[t]
    \centering
    \includegraphics[width=0.93\linewidth]{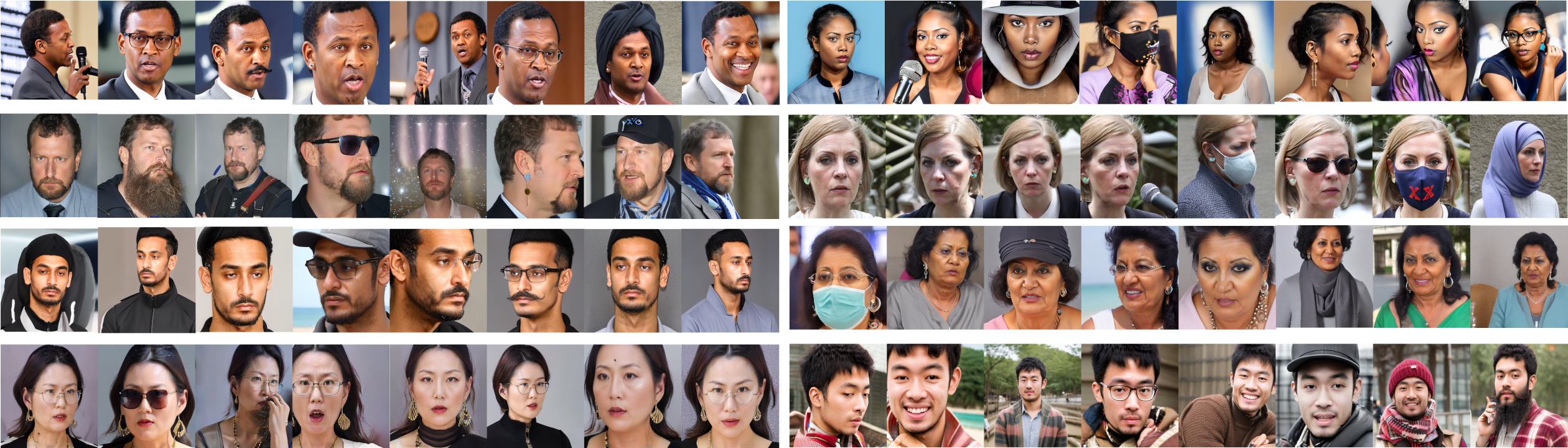}
    \caption{Examples of synthetic identities and variations for different demographic groups using GANDiffFace~\cite{melzi2023gandiffface}.}
    \label{fig:gandiff_ex}
\end{figure*}

Face biometrics is a very popular area in the fields of Computer Vision and Pattern Recognition, finding applications in diverse domains such as person recognition~\cite{wang2021deep, du2022elements}, healthcare~\cite{bisogni2022impact, gomez2021improving}, or e-learning~\cite{daza2023matt}, among others. In the last years, with the rapid evolution of deep learning, we have witnessed a considerable performance improvement in areas such as face recognition (FR)~\cite{deng2019arcface, kim2022adaface}, outperforming the state-of-the-art on established benchmarks. However, FR technology has still room for improvement in several research directions, such as explainability~\cite{deandrestame2024how, crum2023explain}, demographic bias~\cite{terhoerst2021comprehensive, melzi2023synthetic}, privacy~\cite{morales2021sensitivenets, melzi2023multi}, and robustness against challenging conditions~\cite{kim2022adaface}, \textit{e.g.,} aging, pose variations, illumination, occlusions, etc.

Synthetic data has recently appeared as a good solution to mitigate some of these drawbacks, allowing the generation of \textit{i)} a huge number of facial images from different non-existent identities, and \textit{ii)} variability in terms of demographic attributes and scenario conditions. Several approaches have been proposed in the last couple of years for the synthesis of face images, considering state-of-the-art deep learning methods such as Generative Adversarial Networks (GANs)~\cite{qiu2021synface, zhong2021sface}, Diffusion models~\cite{kim2023dcface, boutros2023idiff, kansy2023controllable}, the combination of GAN and Diffusion models~\cite{melzi2023gandiffface}, or alternative methods~\cite{bae2023digiface, zhang2023iti}. Examples of synthetic face images generated using GANDiffFace are shown in Figure~\ref{fig:gandiff_ex}.

However, beyond the generation of novel and realistic synthetic faces, a critical aspect lies in the possible application and benefits of synthetic data to better train FR technology. Recent preliminary studies in the literature have shown the existence of a performance gap between FR systems trained solely on synthetic data and those trained on real data~\cite{kim2023dcface, qiu2021synface}. Nevertheless, the results achieved in the 1$^\text{st}$ edition of the Face Recognition Challenge in the Era of Synthetic Data (FRCSyn) demonstrate the importance of using synthetic data by itself or in combination with real data to mitigate challenges in FR such as demographic bias~\cite{melzi2024frcsyn, melzi2024frcsyna}. It is important to highlight that in the 1$^\text{st}$ edition of the FRCSyn Challenge, only synthetic data from DCFace~\cite{kim2023dcface} and GANDiffFace~\cite{melzi2023gandiffface} methods was allowed to train FR systems. In addition to novel generative methods, another possible improvement of the FR technology could be related to the specific design and training process, taking into account the domain gap between real and synthetic data in some scenarios. For example, we observed in the 1$^\text{st}$ edition of the FRCSyn Challenge that the majority of the teams considered the same deep learning architectures (\textit{e.g.,} ResNet-100~\cite{he2016deep}) and loss functions (\textit{e.g.,} AdaFace~\cite{kim2022adaface}), popularly considered in FR systems trained with real data. 

To promote the proposal of novel face generative methods and the creation of face synthetic databases, as well as specific approaches to better train FR systems with synthetic data, we have organized the 2$^\text{nd}$ edition of the FRCSyn Challenge as part of CVPR 2024\footnote{\href{https://frcsyn.github.io/CVPR2024.html}{https://frcsyn.github.io/CVPR2024.html}}. In this 2$^\text{nd}$ edition, we introduce new sub-tasks enabling participants to train FR systems utilizing synthetic data obtained with the generative frameworks of their choice, offering more freedom compared to the 1$^\text{st}$ edition~\cite{melzi2024frcsyna, melzi2024frcsyn}. In addition, we also consider new sub-tasks featured with different experimental settings, to investigate how FR systems can be trained in both constrained and unconstrained scenarios concerning the amount of synthetic training data. The FRCSyn Challenge aims to answer the following research questions: 
\begin{enumerate}
    \item What are the limits of FR technology trained only with synthetic data?
    \item Can the use of synthetic data be beneficial to reduce the current limitations in FR technology?
\end{enumerate}
These research questions have gained significant importance, particularly after the discontinuation of popular real FR databases due to privacy concerns\footnote{\href{https://exposing.ai/about/news/}{https://exposing.ai/about/news/} \textit{(March, 2024)}} and the introduction of new regulatory laws\footnote{\href{https://artificialintelligenceact.eu}{https://artificialintelligenceact.eu} \textit{(March, 2024)}}. 

The remainder of the paper is organized as follows. Section~\ref{sec:database} focuses on the databases considered in this 2$^\text{nd}$ edition. Section~\ref{sec:frcsyn} explains the experimental setup of the challenge, including the different tasks and sub-tasks, the experimental protocol, metrics,  and restrictions. In Section~\ref{sec:description}, we describe the approaches proposed by the top-6 participating teams. Section~\ref{sec:results} presents the best results achieved in the different tasks and sub-tasks of the 2$^\text{nd}$ edition, emphasizing the key results of the challenge. Finally, in Section \ref{sec:conclusion}, we provide some conclusions, highlighting potential future research directions in the field.


\section{FRCSyn Challenge: Databases} \label{sec:database}

\subsection{Synthetic Databases} \label{sec:syn}
One of the main novelties of this 2$^\text{nd}$ edition of the FRCSyn Challenge is the absence of restrictions on the generative methods allowed to create synthetic data, unlike the 1$^\text{st}$ edition in which only synthetic data created using DCFace~\cite{kim2023dcface} and GANDiffFace~\cite{melzi2023gandiffface} methods was allowed. As a reference, after the registration in the challenge, we provided all the participants with a list of possible state-of-the-art generative frameworks, including DCFace~\cite{kim2023dcface}, GANDiffFace~\cite{melzi2023gandiffface}, DigiFace-1M~\cite{bae2023digiface}, IDiff-Face~\cite{boutros2023idiff}, ID3PM~\cite{kansy2023controllable}, SFace~\cite{zhong2021sface}, SYNFace~\cite{qiu2021synface}, and ITI-GEN~\cite{zhang2023iti}. In addition, we also motivate participants to propose novel face generative methods. In the 2$^\text{nd}$ edition of the FRCSyn Challenge, synthetic data is exclusively utilized in the training stage of FR technology, replicating realistic operational scenarios.

\subsection{Real Databases}\label{sec:realddbb}
For the training of the FR systems (depending on the sub-task, please see Section~\ref{sec:task} for more details), participants are allowed to use only \textbf{CASIA-WebFace}~\cite{yi2014learning}. This database contains $494,414$ face images of $10,575$ real identities collected from the web.

For the final evaluation of the proposed FR systems, we use the same four real databases of the 1$^\text{st}$ edition of the FRCSyn Challenge~\cite{melzi2024frcsyn, melzi2024frcsyna}: \textit{i)} \textbf{BUPT-BalancedFace}~\cite{wang2020mitigating}, designed to address performance disparities across different ethnic groups; \textit{ii)} \textbf{AgeDB}~\cite{moschoglou2017agedb}, including facial images of the same subjects at different ages; \textit{iii)} \textbf{CFP-FP}~\cite{sengupta2016frontal}, presenting facial images from subjects with great changes in pose, including both frontal and profile images; and \textit{iv)} \textbf{ROF}~\cite{erakιn2021recognizing}, consisting of occluded faces with both upper and lower face occlusions.

\section{FRCSyn Challenge: Setup} \label{sec:frcsyn}

\subsection{Tasks}\label{sec:task}

Similar to the 1$^\text{st}$ edition~\cite{melzi2024frcsyna,melzi2024frcsyn}, the challenge has been hosted on Codalab\footnote{\href{https://codalab.lisn.upsaclay.fr/competitions/16970}{https://codalab.lisn.upsaclay.fr/competitions/16970}}. In this 2$^\text{nd}$ edition we also explore the application of synthetic data into the training of FR systems, with a specific focus on addressing two critical aspects in current FR technology: \textit{i)} mitigating demographic bias, and \textit{ii)} enhancing overall performance under challenging conditions that include variations in age and pose, the presence of occlusions, and diverse demographic groups. To investigate these two areas, we propose two distinct tasks, each comprising three sub-tasks considering different types (real/synthetic) and amounts of data for training the FR systems. Consequently, the 2$^\text{nd}$ edition of the FRCSyn Challenge comprises 6 different sub-tasks. In Table~\ref{tab:2}, we summarize the key aspects of the experimental protocol, metrics, and restrictions for each sub-task. 

\textbf{Task 1:} The first proposed task focuses on the use of synthetic data to mitigate demographic biases in FR systems. To assess the effectiveness of the proposed systems, we generate lists of mated and non-mated comparisons using subjects from the BUPT-BalancedFace database \cite{wang2020mitigating}. We take into account eight demographic groups obtained from the combination of four ethnic groups (White, Black, Asian, and Indian) and two genders (Male and Female), and keep these groups balanced in the number of comparisons. In the case of non-mated comparisons, we only consider pairs of subjects within the same demographic group, as these hold greater relevance than non-mated comparisons involving subjects from different demographic groups.

\textbf{Task 2:} The second proposed task focuses on utilizing synthetic data to enhance the overall performance of FR systems under challenging conditions. To assess the effectiveness of the proposed systems, we utilize lists of mated and non-mated comparisons selected from subjects from the different evaluation databases, each one designed to address specific challenges in FR. Specifically, BUPT-BalancedFace is used to consider diverse demographic groups, whereas AgeDB, CFP-FP, and ROF to assess age, pose, and occlusion, respectively.

\begin{table}[t]
    \centering
    \resizebox{0.49\textwidth}{!}{\begin{tabular}{|l|} \hline
         \textbf{Task 1:} synthetic data for \textbf{demographic bias mitigation} \\
         \quad \textbf{Baseline}: training with only CASIA-WebFace \cite{yi2014learning}. \\
         \quad \textbf{Metrics}: accuracy (for each demographic group).\\
         \quad \textbf{Ranking}: average vs SD accuracy, see Section \ref{sec:metrics} for more details. \\ \hline
         
         \textbf{\textcolor{blue}{Sub-Task 1.1: [\textcolor{mygreen}{constrained}]}} training exclusively with \textbf{synthetic} data \\
         \quad \textbf{Train}: maximum $500K$ face images (\textit{e.g.,} $10K$ identities and $50$ images per identity). \\
         \quad \textbf{Eval}: BUPT-BalancedFace \cite{wang2020mitigating}. \\ \hline
         
         \textbf{\textcolor{blue}{Sub-Task 1.2: [\textcolor{mygreen}{unconstrained}]}} training exclusively with \textbf{synthetic} data \\
         \quad \textbf{Train}: no restrictions in terms of the number of face images. \\
         \quad \textbf{Eval}: BUPT-BalancedFace. \\ \hline
         
         \textbf{\textcolor{blue}{Sub-Task 1.3: [\textcolor{mygreen}{constrained}]}} training with \textbf{real and synthetic} data \\ 
         \quad \textbf{Train}: CASIA-WebFace, and maximum $500K$ face synthetic images. \\ 
         \quad \textbf{Eval}: BUPT-BalancedFace. \\ \hline \hline
         
         \textbf{Task 2:} synthetic data for \textbf{overall performance improvement} \\
         \quad \textbf{Baseline}: training with only CASIA-WebFace. \\
         \quad \textbf{Metrics}: accuracy (for each evaluation database).\\
         \quad \textbf{Ranking}: average accuracy, see Section \ref{sec:metrics} for more details.\\ \hline
         
         \textbf{\textcolor{blue}{Sub-Task 2.1: [\textcolor{mygreen}{constrained}]}} training with only \textbf{synthetic} data \\ 
         \quad \textbf{Train}: maximum $500K$ face images.\\
         \quad \textbf{Eval}: BUPT-BalancedFace, AgeDB~\cite{moschoglou2017agedb}, CFP-FP~\cite{sengupta2016frontal}, and ROF~\cite{erakιn2021recognizing}. \\ \hline

         \textbf{\textcolor{blue}{Sub-Task 2.2: [\textcolor{mygreen}{unconstrained}]}} training with only \textbf{synthetic} data \\ 
         \quad \textbf{Train}: no restrictions in terms of the number of face images. \\
         \quad \textbf{Eval}: BUPT-BalancedFace, AgeDB, CFP-FP, and ROF. \\  \hline
         
         \textbf{\textcolor{blue}{Sub-Task 2.3: [\textcolor{mygreen}{constrained}]}} training with \textbf{real and synthetic} data \\
         \quad \textbf{Train}: CASIA-WebFace, and maximum $500K$ face synthetic images. \\
         \quad \textbf{Eval}: BUPT-BalancedFace, AgeDB, CFP-FP, and ROF. \\ \hline
    \end{tabular}}
    \caption{Tasks and sub-tasks for the 2$^\text{nd}$ FRCSyn Challenge and their respective metrics and databases. SD = Standard Deviation.}
    \label{tab:2}
\end{table}

\subsection{Experimental protocol}
\textbf{Training:} The 6 sub-tasks introduced in the 2$^\text{nd}$ edition of the FRCSyn Challenge are mutually independent. This implies that participants have the flexibility to participate in any number of sub-tasks based on their preferences. For each selected sub-task, participants are required to develop and train the same FR system twice: \textit{i)} using the authorized real database exclusively, i.e. CASIA-WebFace \cite{yi2014learning}, and \textit{ii)} following the specific requirements of the chosen sub-task, as summarized in Table \ref{tab:2}. According to this protocol, participants must provide both the \textit{baseline system} and the \textit{proposed system} for the specific sub-task. The baseline system plays a critical role in evaluating the impact of synthetic data on training and serves as a reference point for comparing against the conventional practice of training solely with real databases. To maintain consistency, the baseline FR system, trained exclusively with real data, and the proposed FR system, trained according to the specifications of the selected sub-task, must have the same architecture.

\textbf{Evaluation:} In each sub-task, participants received the comparison files comprising both mated and non-mated comparisons, which are used to evaluate the performance of their proposed FR systems. Task 1 involves a single comparison file containing balanced comparisons of different demographic groups of the BUPT~\cite{wang2020mitigating} database, while Task 2 comprises four comparison files, each corresponding to each of the specific real-world databases considered (i.e., BUPT, AgeDB~\cite{moschoglou2017agedb}, CFP-FP~\cite{sengupta2016frontal}, and ROF~\cite{erakιn2021recognizing}). During the evaluation of each sub-task, participants are required to submit via Codalab three files per database: \textit {i)} the scores of the baseline system, \textit{ii)} the scores of the proposed system, and \textit{iii)} the decision threshold for each FR system \textit{(i.e.,} baseline and proposed). The submitted scores must fall within the range of $[0, 1]$, with lower scores indicating non-mated comparisons, and vice versa.

\subsection{Evaluation Metrics}\label{sec:metrics}
We evaluate FR systems using a protocol based on lists of mated and non-mated comparisons for each sub-task and database. From the scores and thresholds provided by participants, we calculate the binary decision and the verification accuracy. Additionally, we calculate the gap to real (GAP) \cite{kim2023dcface} as follows: $\text{GAP} = \left( \text{REAL} - \text{SYN} \right)/\text{SYN}$, with $\text{REAL}$ representing the verification accuracy of the baseline system and $\text{SYN}$ the verification accuracy of the proposed system, trained with synthetic (or real + synthetic) data. Other metrics such as False Non-Match Rate (FNMR) at 1\% False Match Rate (FMR), which are very popular for the analysis of FR systems in real-world applications, can also be computed from the scores provided by participants. Due to the lack of space, comprehensive evaluations of the proposed systems will be conducted in subsequent studies, including FNMRs and metrics for each demographic group and database used for evaluation. Next, we explain how participants are ranked in the different tasks.

\textbf{Task 1:} To rank participants and determine the winners of Sub-Tasks 1.1, 1.2, and 1.3, we closely examine the trade-off between the average (AVG) and standard deviation (SD) of the verification accuracy across the eight demographic groups defined in Section \ref{sec:task}. We define the trade-off metric (TO) as follows: $\text{TO} = \text{AVG} - \text{SD}$. This metric corresponds to plotting the average accuracy on the x-axis and the standard deviation on the y-axis in 2D space. We draw multiple 45-degree parallel lines to find the winning team whose performance falls to the far right side of these lines. With this proposed metric, we reward FR systems that achieve good levels of performance and fairness simultaneously, unlike common benchmarks based only on recognition performance. The standard deviation of verification accuracy across demographic groups is a common metric for assessing bias and should be reported by any work addressing demographic bias mitigation.

\textbf{Task 2:} To rank participants and determine the winners of Sub-Tasks 2.1, 2.2, and 2.3, we consider the average verification accuracy across the four databases used for evaluation, described in Section \ref{sec:task}. This approach allows us to evaluate four challenging aspects of FR simultaneously: \textit{i)} diverse demographic groups, \textit{ii)} pose variations, \textit{iii)} aging, and \textit{iv)} presence of occlusions providing a comprehensive evaluation of FR systems in real operational scenarios.

\subsection{Restrictions}
Regarding the FR system, participants have the freedom to choose any architecture for each sub-task, provided that the system's number of Floating Point Operations Per Second (FLOPs) does not exceed 50 GFLOPs. This threshold has been established to facilitate the exploration of innovative architectures and encourage the use of diverse models while preventing the dominance of excessively large models. Participants are also free to utilize their preferred training modality, with the requirement that only the specified databases are used for training. This means that no additional databases can be employed during the training phase, such as to adapt the verification thresholds. Participants are allowed to use non-face databases for pre-training purposes and employ traditional data augmentation techniques using the authorized training databases. Regarding the synthetic data used to train the FR system in each sub-task, we allow participants to use any existing/novel database and face generative framework, regardless of how the model is trained.

To maintain the integrity of the evaluation process, the organizers reserve the right to disqualify participants if anomalous results are detected or if participants fail to adhere to the challenge's rules.

\section{FRCSyn Challenge: Systems Description} \label{sec:description}
The 2$^\text{nd}$ edition of the FRCSyn Challenge received significant interest, with $78$ international teams correctly registered, comprising research groups from both industry and academia. These teams work in various domains, including FR, generative AI, and other aspects of computer vision, such as demographic fairness and domain adaptation. Finally, $23$ teams submitted their scores, receiving all sub-tasks great attention. The submitting teams are geographically distributed, with fourteen teams from Asia, six teams from Europe, and three teams from America. Table \ref{tab:best_teams} provides a general overview of the teams that ranked among the top-6 in at least one sub-task, including the sub-tasks in which they participated. Next, we describe briefly the approaches proposed for each team. 

\begin{table}[t]
    \centering
    \resizebox{0.47\textwidth}{!}{\begin{tabular}{lclc}
    \textbf{Team} & \textbf{Affiliations} & \textbf{Country} & \textbf{Sub-Tasks} \\ \hline
    ADMIS & $4, 5$ & China & all \\
    OPDAI & $6$ & China & all \\
    ID R\&D & $7$ & USA & all \\
    K-IBS-DS & $8, 9$ & South Korea & all \\
    CTAI & $10$ & China & all \\
    Idiap-SynthDistill & $11, 12, 13$ & Switzerland & 1.2 - 2.2 \\
    INESC-IGD & $14, 15, 16$ & Portugal and Germany & all \\
    UNICA-IGD-LSI & $16, 17, 18$ & Italy, Germany, Slovenia & all \\
    SRCN\_AIVL & $19, 20, 21, 22$ & China & 1.1 \\
    CBSR-Samsung & $19, 21, 22$ & China & 1.3 - 2.3 \\
    BOVIFOCR-UFPR & $23, 24, 25$ & Brazil & 1.2 - 2.1 \\ \hline
    \end{tabular}}
    \caption{Description of the teams that ranked among the top-6 in at least one sub-task, ordered by the average rank in all the sub-tasks. The numbers reported in the column `affiliations' refer to the ones provided in the title page.}
    \label{tab:best_teams}
\end{table}

\paragraph{ADMIS (All sub-tasks):}
They used an IDiff-Face-based~\cite{boutros2023idiff} Latent Diffusion Model (LDM) to synthesize face images. Specifically, they trained an identity-conditioned LDM using ID embeddings extracted from CASIA-WebFace~\cite{yi2014learning} with a pre-trained ElasticFace~\cite{boutros2022elasticface} IResNet-101~\cite{duta2021improved} model. As the LDM takes the ID embeddings as context, they employed an unconditional Denoising Diffusion Probabilistic Model (DDPM) trained on the FFHQ database~\cite{karras2019style} as a context generator. This produced $400K$ images, from which they extracted approximately $30K$ unique ID embeddings with a $0.3$ similarity threshold using the pre-trained ElasticFace model, creating a context database. Furthermore, they accelerated the sampling process of the LDM using a DDIM~\cite{song2021denoising}. For the training of the FR model, they generated $49$ images for each context. They adopted the ID oversampling strategy from DCFace~\cite{kim2023dcface} and performed it five times for each ID to enhance consistency. As a result, $10K$ contexts were utilized for Sub-Tasks 1.1 and 2.1, while $30K$ for Sub-Tasks 1.2 and 2.2. For Sub-Tasks 1.3 and 2.3, they expanded Sub-Tasks 1.1 and 2.1 with the CASIA-WebFace database. They applied the ArcFace~\cite{deng2019arcface} loss and random cropping augmentation during training. Both the baseline and proposed models used IResNet-101 architectures~\cite{duta2021improved}. 

\noindent Code: \footnotesize\href{https://github.com/zzzweakman/CVPR24_FRCSyn_ADMIS}{https://github.com/zzzweakman/CVPR24\_FRCSyn\_ADMIS}\normalsize

\paragraph{OPDAI (All sub-tasks):}
They initially used the data provided by DCFace~\cite{kim2023dcface}, generating then $10$ more face images for each ID with large pose variations and occlusions using Photomaker~\cite{li2024photomaker}. They randomly replaced these images in the original DCFace data to ensure that the total number of samples meets the requirement of $500K$. During the Photomaker inference, they adopted a batch size of $1$ and used random prompts including age, pose, and image quality to ensure the diversity of the generated samples. For Sub-Tasks 1.2 and 2.2, they combined this data with the $1.2M$ version of DCFace, while for Sub-Tasks 1.3 and 2.3, it was merged with CASIA-WebFace~\cite{yi2014learning}. For Sub-Tasks 1.2, 1.3, 2.2, and 2.3 they did not merge nor denoise samples from different databases, following the Partial FC approach~\cite{an2021partial}. Also, they obtained the loss of different databases in independent AdaFace~\cite{kim2022adaface} heads, calculating the final loss as the average of the multiple heads. Both baseline and proposed models are based on IResNet-100~\cite{duta2021improved} architectures, with horizontal flipping. 

\noindent Code: \footnotesize\href{https://github.com/mightycatty/frcsyn_cvpr2024.git}{https://github.com/mightycatty/frcsyn\_cvpr2024.git}\normalsize

\paragraph{ID R\&D (All sub-tasks):}
To generate the synthetic data, they used two models trained on WebFace42M~\cite{zhu2021webface260m}, one based on Hourglass Diffusion Transformers~\cite{crowson2024scalable} and the other on StyleNAT~\cite{walton2022stylenat}, enhanced with a FR model~\cite{smirnov2022prototype}. They used classifier weights of the trained Prototype Memory~\cite{smirnov2022prototype} to get $50K$ identity vectors, of which $20K$ were randomly selected and $30K$ were uniformly sampled from the $1K$ clusters obtained using k-means, to get demographic diversity. For each identity, they generated $5$ images using each of the two generative models. This data was used to train IResNet-200~\cite{duta2021improved} with UniFace~\cite{zhou2023uniface} loss for $28$ epochs. One network was trained with color, geometric augmentations, and FaceMix-B~\cite{garaev2023facemixa}, and the other network used only random horizontal flipping. These two networks were combined in an ``ensemble", where the first one received the original image, and the second one a mirrored copy. They used the same model for Sub-Tasks 1.1, 1.2, 2.1 and 2.2. For Sub-Tasks 1.3 and 2.3, they combined the synthetic data and CASIA-WebFace, training two models, one on the mixed data, and the other on the CASIA-WebFace.

\paragraph{K-IBS-DS (All sub-tasks):}
Inspired by SlackedFace~\cite{low2023slackedface}, they made two modifications to enhance the AdaFace~\cite{kim2022adaface} FR classifier. First, they made a more reliable weight initialization for uniformity across identity prototypes in the unit sphere and replaced the L2-norm with the face recognizability index from~\cite{low2023slackedface}. Regarding the synthetic data, they used DCFace~\cite{kim2023dcface} with $500K$ and $1.2M$ face images (depending on the sub-task). The training stage was in line with~\cite{kim2023dcface} and~\cite{bae2023digiface}, including optimizer, learning rate, etc. For Sub-Tasks 1.3 and 2.3, the first $10K$ subjects of the CASIA-WebFace~\cite{yi2014learning} were assigned for training, and the remaining ones for performance validation using random pairs with challenging conditions (identified based on the poorest L2-norm values~\cite{melzi2023gandiffface}). The final score is obtained by aggregating the comparison scores of ResNet with Squeeze-and-Excitation (SE) blocks~\cite{hu2018squeeze} models of $50$, $100$, and $152$ layers, along with the horizontally flipped instances through score fusion. 

\noindent Code: \footnotesize\href{https://github.com/kalebmes/cvpr_frcsyn}{https://github.com/kalebmes/cvpr\_frcsyn}\normalsize

\paragraph{CTAI (All sub-tasks):}
By analyzing popular synthetic data, they found that intra-class and inter-class noise was widely present. Data cleaning can effectively remove the bad examples of synthetic data and retain important images from a large amount of synthetic data. In order to select the optimal synthetic data, they first trained an IResNet-100~\cite{duta2021improved} model with Squeeze-and-Excitation (SE)~\cite{hu2018squeeze} blocks using CASIA-WebFace~\cite{yi2014learning} to extract features of synthetic images from DCFace~\cite{kim2023dcface}, GANDiffFace~\cite{melzi2023gandiffface}, and DigiFace~\cite{bae2023digiface}. Subsequently, they used DBSCAN clustering to segregate intra-class noise and removed IDs with a class center feature cosine similarity greater than $0.5$. Finally, they used the cleaned synthetic data merged with CASIA-WebFace to finetune the IResNet-100 for a second data refinement. From the final refined synthetic dataset, they sampled $500K$ face images while retaining as many IDs as possible to build their synthetic training set. Regarding the FR model, in particular Sub-Task 2.3 in which they achieved their highest position among all sub-tasks, they trained IResNet-100 with AdaFace~\cite{kim2022adaface} loss (A1) and CosFace~\cite{wang2018cosface} loss (A2) with mask and occlusion augmentation on CASIA-WebFace and the refined synthetic data. They used an ensemble of A1, A2, and a model trained with only synthetic data. Furthermore, data augmentation was employed to enhance all features. 

\noindent Code: \footnotesize\href{https://github.com/liuhao-lh/FRCSyn-Challenge}{https://github.com/liuhao-lh/FRCSyn-Challenge}\normalsize

\paragraph{Idiap-SynthDistill (Sub-Tasks 1.2 and 2.2):}
The proposed method was based on SynthDistill~\cite{shahreza2023synthdistill}, which is an end-to-end approach, generating synthetic images and training the FR model in the same training loop. Instead of using the pre-trained model in a separate step, they directly used it in the training loop for supervision, while a new student FR model was trained fully using synthetic data generated from a StyleGAN model. For generating synthetic images, they trained StyleGAN2~\cite{karras2020analyzing} with the CASIA-WebFace database~\cite{yi2014learning} and then dynamically generated synthetic images during training based on the training loss. For the dynamic image generation, they used the training loss to find the most difficult synthetic image in each batch, and then they generated a new batch of synthetic images by re-sampling the most difficult samples. Regarding the FR model, they used a model with the IResNet-101~\cite{duta2021improved} architecture and trained it with synthetic data using SynthDistill. They used the Adam optimizer with an initial learning rate of $0.001$ and trained their student model with the same loss function as in~\cite{shahreza2023synthdistill}. For thresholding, a subset of DCFace~\cite{kim2023dcface} was used to determine the optimal threshold for maximizing verification accuracy, using a 10-fold cross-validation approach based on a random selection of identities and comparison pairs. 

\noindent Code: \footnotesize\href{https://gitlab.idiap.ch/bob/bob.paper.ijcb2023_synthdistill}{https://gitlab.idiap.ch/bob/bob.paper.ijcb2023\_synthdistill}\normalsize

\paragraph{INESC-IGD (All sub-tasks):}
In all sub-tasks they trained a ResNet-100 with ElasticCosFac-Plus loss~\cite{boutros2022elasticface} using the settings presented in~\cite{boutros2022elasticface}. For the training dataset, DCFace~\cite{kim2023dcface}, IDiff-Face Uniform, and IDiff-Face Two-stage~\cite{boutros2023idiff} datasets were merged and their images were labeled with ethnicity labels using a similar approach to~\cite{neto2023compressed}. 
For Sub-Tasks 1.1 and 2.1, they created a synthetic training dataset containing $500K$ face images by sampling $7K$ balanced identities, in terms of ethnicity labels. For Sub-Tasks 1.2 and 2.2, they created a synthetic training dataset containing $2.1M$ face images by sampling $50K$ identities from the training datasets. For Sub-Tasks 1.3 and 2.3, two instances of ResNet-100 were trained on CASIA-WebFace and a subset of synthetic datasets ($400K$ images of $9K$ identities), respectively. The synthetic datasets were sampled from DCFace and IDiff-Face. During the testing phase of Sub-Tasks 1.3 and 2.3, feature embeddings were obtained from trained models and the weighted sum of $0.5$ score-level fusion was utilized. During the FR training of all sub-tasks, the training datasets were augmented using the RandAug utilized in IDiff-Face and occluded augmentation~\cite{neto2022ocfr} with probabilities of $0.4$.

\noindent Code: \footnotesize\href{https://github.com/NetoPedro/Equilibrium-Face-Recognition}{https://github.com/NetoPedro/Equilibrium-Face-Recognition}\normalsize

\paragraph{UNICA-IGD-LSI (All sub-tasks):}
They used the DCFace~\cite{kim2023dcface} synthetic dataset as it led to remarkable performance gains under well-known evaluation benchmarks for face verification, while combined with real data~\cite{atzori2024if}. They trained a ResNet-100~\cite{he2016deep} network using CosFace loss~\cite{wang2018cosface} with a margin penalty of 0.35 and a scale term of 64. The similarity mean difference between real-only and synthetic-only samples was scaled and added to the loss value. They trained the model for $40$ epochs with a batch size of $512$ and an initial learning rate of $0.1$, which was divided by $10$ after $10$, $22$, $30$, and $40$ epochs. During the training phase, the synthetic samples were augmented using RandAugment with $4$ operations and a magnitude of 16, following~\cite{boutros2023unsupervised, atzori2024if}. For Sub-Tasks 1.3 and 2.3, the chosen synthetic dataset was combined with CASIA-Webface~\cite{yi2014learning}, obtaining a total of $1M$ images from $20,572$ identities. 

\noindent Code: \footnotesize\href{https://github.com/atzoriandrea/FRCSyn2}{https://github.com/atzoriandrea/FRCSyn2}\normalsize

\paragraph{SRCN\_AIVL (Sub-Task 1.1):}
They selected $400K$ samples from the DCFace~\cite{kim2023dcface} database and labeled the ethnicity of each subject, as they considered that the racial distribution gap may lead to bad performance in testing. Based on this insight, they trained IDiff-Face~\cite{boutros2023idiff} with CASIA-WebFace~\cite{yi2014learning} database generating $100K$ synthetic face images of specific races. Regarding the FR system, they used two custom ResNet-101~\cite{he2016deep} trained with AdaFace loss~\cite{kim2022adaface} function. The models were trained for $60$ epochs with an initial learning rate of $0.1$, which was adjusted at predefined milestones. Their training data underwent further preprocessing, including padding crop augmentation, low-resolution augmentation, photometric augmentation, random grayscale, and normalization. For the inference, data preprocessing involved an MTCNN~\cite{zhang2016joint} and resizing all data. After cropping and alignment, they fed the image and the flipped image into the two models. After obtaining the two feature embeddings, they combined them and performed the similarity calculation with these embeddings. 

\noindent Code: \footnotesize\href{https://github.com/Value-Jack/2nd-Edition-FRCSyn}{https://github.com/Value-Jack/2nd-Edition-FRCSyn}\normalsize

\paragraph{CBSR-Samsung (Sub-Tasks 1.3 and 2.3):}
They first trained a FR model using CASIA-WebFace~\cite{yi2014learning}. Then, they used it to de-overlap DCFace~\cite{kim2023dcface} from CASIA, as DCFace was trained using that real database. For the synthetic dataset, they compared the performance of models trained with three synthetic datasets, including GANDiffFace~\cite{melzi2023gandiffface}, DCFace, and IDiffFace~\cite{boutros2023idiff}, and finally selected DCFace as the only synthetic training set. They created a validation dataset including three subsets for three different testing scenarios: \textit{i)} random sample pairs from DCFace; \textit{ii)} randomly positioned vertical bar masks to the images to simulate the self-occlusion due to pose; and \textit{iii)} add a mask and sunglasses to images by detecting the landmarks~\cite{wang2021facex}. All validation subsets consist of $6K$ positive pairs and $6K$ negative pairs. Finally, they concatenated these subsets as the validation set. Subsequently, they conducted an intra-class clustering for all datasets using DBSCAN ($0.3$ threshold) and removed the samples that were separated from the class center. They merged the refined datasets and trained IResNet-100~\cite{duta2021improved} with AdaFace loss~\cite{kim2022adaface}. In addition, they adopted two augmentation strategies, \textit{i.e.,} photometric augmentation and rescaling. After that, they trained two FR models using occlusion augmentation with $10\%$ and $30\%$ probability, respectively. Finally, they submitted the average similarity score of the two models.

\paragraph{BOVIFOCR-UFPR (Sub-Tasks 1.2 and 2.1):}
They chose DCFace~\cite{kim2023dcface} as the synthetic dataset and ResNet-100~\cite{he2016deep} as the backbone, trained with the ArcFace~\cite{deng2019arcface} loss function. The images used for training were augmented using a Random Flip with a probability of $0.5$. They also applied random erasing and RandAugment as additional augmentations. The model was trained using the Insightface library for $20$ epochs within a batch size of $128$, running for approximately $78K$ iterations. 

\noindent Code: \footnotesize\href{https://github.com/PedroBVidal/insightface}{https://github.com/PedroBVidal/insightface}\normalsize

\begin{table}[t]
\footnotesize
\centering
\resizebox{0.47\textwidth}{!}{\begin{tabular}{clcccc}
\multicolumn{6}{c}{\textbf{Sub-Task 1.1 (Bias Mitigation): Synthetic Data (Constrained)}} \\ \hline
\textbf{Pos.} & \textbf{Team} & \textbf{TO {[}\%{]}} & \textbf{AVG {[}\%{]}} & \textbf{SD {[}\%{]}} & \textbf{GAP {[}\%{]}} \\ \hline
1 & ID R\&D & 96.73 & 97.55 & 0.82 & -5.31 \\
2 & ADMIS & 94.30 & 95.10 & 0.80 & 1.47 \\
3 & SRCN\_AIVL & 94.06 & 95.12 & 1.07 & -0.54 \\
4 & OPDAI & 93.75 & 94.92 & 1.17 & 1.02 \\
5 & CTAI & 93.21 & 94.74 & 1.53 & -0.63 \\
6 & K-IBS-DS & 92.91 & 94.11 & 1.20 & 1.58 \\ \hline
\end{tabular}}

\bigskip

\resizebox{0.47\textwidth}{!}{\begin{tabular}{clcccc}
\multicolumn{6}{c}{\textbf{Sub-Task 1.2 (Bias Mitigation): Synthetic Data (Unconstrained)}} \\ \hline
\textbf{Pos.} & \textbf{Team} & \textbf{TO {[}\%{]}} & \textbf{AVG {[}\%{]}} & \textbf{SD {[}\%{]}} & \textbf{GAP {[}\%{]}} \\ \hline
1 & ID R\&D & 96.73 & 97.55 & 0.82 & -5.31 \\
2 & ADMIS & 95.72 & 96.50 & 0.78 & -0.56 \\
3 & OPDAI & 94.12 & 95.22 & 1.11 & 0.71 \\
4 & INESC-IGD & 94.05 & 95.22 & 1.17 & 1.04 \\
5 & K-IBS-DS & 93.72 & 94.88 & 1.16 & 0.77 \\
6 & CTAI & 93.21 & 94.74 & 1.53 & -0.63 \\ \hline
\end{tabular}}

\bigskip

\resizebox{0.47\textwidth}{!}{\begin{tabular}{clcccc}
\multicolumn{6}{c}{\textbf{Sub-Task 1.3 (Bias Mitigation): Synthetic + Real Data (Constrained)}} \\ \hline
\textbf{Pos.} & \textbf{Team} & \textbf{TO {[}\%{]}} & \textbf{AVG {[}\%{]}} & \textbf{SD {[}\%{]}} & \textbf{GAP {[}\%{]}} \\ \hline
1 & ADMIS & 96.50 & 97.25 & 0.75 & -1.33 \\
2 & K-IBS-DS & 96.17 & 96.92 & 0.75 & -1.37 \\
3 & UNICA-IGD-LSI & 96.00 & 96.70 & 0.70 & -5.33 \\
4 & OPDAI & 95.96 & 96.80 & 0.84 & -0.03 \\
5 & INESC-IGD & 95.65 & 96.33 & 0.67 & -0.12 \\
6 & CBSR-Samsung & 95.57 & 96.54 & 0.97 & -24.43 \\ \hline
\end{tabular}}

\bigskip

\resizebox{0.47\textwidth}{!}{\begin{tabular}{clcc}
\multicolumn{4}{c}{\textbf{Sub-Task 2.1 (Overall Improvement): Synthetic Data (Constrained)}} \\ \hline
\textbf{Pos.} & \textbf{Team} & \textbf{AVG {[}\%{]}} & \textbf{GAP {[}\%{]}} \\ \hline
1 & OPDAI & 91.93 & 3.09 \\
2 & ID R\&D & 91.86 & 2.99 \\
3 & ADMIS & 91.19 & 2.78 \\
4 & K-IBS-DS & 91.05 & 2.60 \\
5 & CTAI & 90.59 & -1.94 \\
6 & BOVIFOCR-UFPR & 89.97 & 3.71 \\ \hline
\end{tabular}}

\bigskip

\resizebox{0.47\textwidth}{!}{\begin{tabular}{clcc}
\multicolumn{4}{c}{\textbf{Sub-Task 2.2 (Overall Improvement): Synthetic Data (Unconstrained)}} \\ \hline
\textbf{Pos.} & \textbf{Team} & \textbf{AVG {[}\%{]}} & \textbf{GAP {[}\%{]}} \\ \hline
1 & Idiap-SynthDistill & 93.50 & -0.05 \\
2 & ADMIS & 92.92 & 0.21 \\
3 & OPDAI & 92.04 & 3.00 \\
4 & ID R\&D & 91.86 & 2.99 \\
5 & K-IBS-DS & 91.61 & 1.96 \\
6 & CTAI & 90.59 & -1.94 \\ \hline
\end{tabular}}

\bigskip

\resizebox{0.47\textwidth}{!}{\begin{tabular}{clcc}
\multicolumn{4}{c}{\textbf{Sub-Task 2.3 (Overall Improvement): Synthetic + Real Data (Constrained)}} \\ \hline
\textbf{Pos.} & \textbf{Team} & \textbf{AVG {[}\%{]}} & \textbf{GAP {[}\%{]}} \\ \hline
1 & K-IBS-DS & 95.42 & -2.15 \\
2 & OPDAI & 95.23 & -0.52 \\
3 & CTAI & 94.56 & -6.01 \\
4 & CBSR-Samsung & 94.20 & -4.40 \\
5 & ADMIS & 94.15 & -1.10 \\
6 & ID R\&D & 94.05 & 0.07 \\ \hline
\end{tabular}}
\caption{Ranking for the six sub-tasks, according to the metrics described in Section \ref{sec:metrics}. TO = Trade-Off, AVG = Average accuracy, SD = Standard Deviation of accuracy, GAP = Gap to Real.}
\label{tab:results}
\end{table}

\section{FRCSyn Challenge: Results} \label{sec:results}
Table~\ref{tab:results} presents the rankings for the different sub-tasks considered in the 2$^\text{nd}$ edition of the FRCSyn Challenge. In general, the rankings for Sub-Tasks 1.1, 1.2, and 1.3 (bias mitigation), corresponding to the descending order of TO, closely align with the ascending order of SD (\textit{i.e.,} from less to more biased FR systems). Notably, the winner of Sub-Tasks 1.1 and 1.2, ID R\&D (96.73\% TO), exhibits a considerable negative GAP value (-5.31\%), indicating higher accuracy when training the FR system with synthetic data compared to real data (\textit{i.e.,} CASIA-WebFace~\cite{yi2014learning}). Furthermore, when the limitation in the number of synthetic images is removed (\textit{i.e.,} Sub-Task 1.2), the TO value of most FR systems increases, obtaining a performance improvement and fairness simultaneously. For example, for the ADMIS team (top-2) the TO value increases to 95.72\% (\textit{i.e.,} 1.42\% TO general improvement from Sub-Tasks 1.1 to 1.2), with a GAP value of -0.56\%. These results highlight the advantages of synthetic data, including the potential for generating an infinite number of face images to reduce bias in current FR technology. Finally, for completeness, we analyze in Sub-Task 1.3 the case of adding real and synthetic data to the FR training process. In general, we can observe better TO values, in addition to negative GAP values for all the top-6 teams, \textit{e.g.,} ADMIS (96.50\% TO, -1.33 GAP), K-IBS-DS (96.17\% TO, -1.37\% GAP), and UNICA-IGD-LSI (96.00\% TO, -5.33\% GAP). These results prove that the combination of synthetic and real data achieves higher FR performance compared to training only with real data. In addition, it is also interesting to compare the best results achieved in Sub-Task 1.2, \textit{i.e.,} unconstrained synthetic data, and Sub-Task 1.3, \textit{i.e.,} constrained synthetic + real data. The ID R\&D team achieves 96.73\% TO in Sub-Task 1.2 whereas ADMIS achieves 96.50\% TO in Sub-Task 1.3, proving that it is possible to obtain better results using only unlimited synthetic data than including real data.

For Task 2, the average accuracy across databases in the different sub-tasks is lower than the accuracy achieved for BUPT-BalancedFace~\cite{wang2020mitigating} in Task 1, emphasizing the additional challenges introduced by AgeDB~\cite{moschoglou2017agedb}, CFP-FP~\cite{sengupta2016frontal}, and ROF~\cite{erakιn2021recognizing} real databases considered for evaluation. Also, although good results are achieved in Sub-Task 2.1 when training only with synthetic data (\textit{e.g.,} 91.93\% AVG for OPDAI), the positive GAP values provided by most of the top-6 teams are the greatest from all the sub-tasks, indicating that synthetic data alone currently struggles to completely replace real data for training FR systems. Nevertheless, in Sub-Task 2.2 in which there are no restrictions in the number of synthetic images to use, the Idiap-SynthDistill team (top-1) achieves much better results (93.50\% AVG) with a GAP value of -0.05, proving that unlimited synthetic data by itself can even outperform limited real data. Finally, in Sub-Task 2.3, most of the teams report better AVG and higher negative GAP values (\textit{e.g.,} 95.42\% AVG and -2.15\% GAP for the K-IBS-DS team, top-1), which suggests that synthetic data combined with real data can mitigate existing limitations within FR technology.

Finally, analyzing the contributions of all eleven top teams, a notable trend emerges, showing the prevalence of well-established methodologies. ResNet~\cite{he2016deep} or IResNet~\cite{duta2021improved} backbones were chosen by all the teams for their wide adoption in state-of-the-art FR approaches. The AdaFace~\cite{kim2022adaface} and ArcFace \cite{deng2019arcface} loss functions were widely used, featuring in the approaches of most of the teams, except for ID R\&D which used the recent UniFace~\cite{zhou2023uniface} or UNICA which used CosFace~\cite{wang2018cosface}. Notably, all the teams used DCFace~\cite{kim2023dcface} alone or combined with other databases like GANDiffFace~\cite{melzi2023gandiffface}, DigiFace~\cite{bae2023digiface}, or IDiffFace~\cite{boutros2023idiff}, considering also interesting approaches based on synthetic data cleaning and selection for some teams such as CTAI and CBSR-Samsung. ID R\&D and Idiap-SynthDistill were the only teams that used different approaches to generate the synthetic data. In particular, an Hourglass Diffusion Transformer~\cite{crowson2024scalable} and StyleNAT~\cite{walton2022stylenat} by the ID R\&D team, and dynamic image generation using StyleGAN2~\cite{shahreza2023synthdistill} by the Idiap-SynthDistill team.

\section{Conclusion} \label{sec:conclusion}
The 2$^\text{nd}$ edition of the FRCSyn Challenge has presented a comprehensive exploration of the applications of synthetic data in FR, effectively addressing existing limitations in the field. In this 2$^\text{nd}$ edition, two additional sub-tasks have been introduced, showing that impressive results can be achieved using unlimited synthetic data, even outperforming in some cases the scenario of training with real data. With an increased number of participants in this edition, we have witnessed a considerable performance improvement in all sub-tasks in comparison to the 1$^\text{st}$ edition~\cite{melzi2024frcsyn, melzi2024frcsyna}. This has been possible thanks to the proposal of novel methods to generate and select better synthetic data, as well as FR models and loss functions. These approaches can be compared across a variety of sub-tasks, with many being reproducible thanks to the materials made available by the participating teams. Future works will be oriented to a more detailed analysis of the results and comparison with recent challenges in the topic, such as SDFR~\cite{shahreza2024sdfr}. We also plan to transform the CodaLab platform into an ongoing challenge, similar to what we did in the 1$^\text{st}$ edition~\cite{melzi2024frcsyn}.

\section*{Acknowledgements}
{\footnotesize This study has received funding from the European Union's Horizon 2020 TReSPAsS-ETN (No 860813) and is supported by INTER-ACTION (PID2021-126521OB-I00 MICINN/FEDER), Cátedra ENIA UAM-VERIDAS en IA Responsable (NextGenerationEU PRTR TSI-100927-2023-2) and R\&D Agreement DGGC/ UAM/FUAM for Biometrics and Cybersecurity.
It is also supported by the German Federal Ministry of Education and Research and the Hessian Ministry of Higher Education, Research, Science and the Arts within their joint support of the National Research Center for Applied Cybersecurity ATHENE.
K-IBS-DS was supported by the Institute for Basic Science, Republic of Korea (IBS-R029-C2). UNICA-IGD-LSI was supported by the ARIS program P2-0250B.}
{
    \small
    \bibliographystyle{ieeenat_fullname}
    \bibliography{main}
}

\end{document}

%% file: preamble.tex
\usepackage[dvipsnames]{xcolor}


%% file: main.bbl
\begin{thebibliography}{53}
\providecommand{\natexlab}[1]{#1}
\providecommand{\url}[1]{\texttt{#1}}
\expandafter\ifx\csname urlstyle\endcsname\relax
  \providecommand{\doi}[1]{doi: #1}\else
  \providecommand{\doi}{doi: \begingroup \urlstyle{rm}\Url}\fi

\bibitem[An et~al.(2021)An, Zhu, Gao, Xiao, Zhao, Feng, Wu, Qin, Zhang, Zhang, and Fu]{an2021partial}
Xiang An, Xuhan Zhu, Yuan Gao, Yang Xiao, Yongle Zhao, Ziyong Feng, Lan Wu, Bin Qin, Ming Zhang, Debing Zhang, and Ying Fu.
\newblock {Partial FC: Training 10 Million Identities on a Single Machine}.
\newblock In \emph{Proc. IEEE/CVF International Conference on Computer Vision Workshops}, 2021.

\bibitem[Atzori et~al.(2024)Atzori, Boutros, Damer, Fenu, and Marras]{atzori2024if}
Andrea Atzori, Fadi Boutros, Naser Damer, Gianni Fenu, and Mirko Marras.
\newblock {If It's Not Enough, Make It So: Reducing Authentic Data Demand in Face Recognition through Synthetic Faces}.
\newblock In \emph{Proc. 18th International Conference on Automatic Face and Gesture Recognition}, 2024.

\bibitem[Bae et~al.(2023)Bae, de~La~Gorce, Baltru\v{s}aitis, Hewitt, Chen, Valentin, Cipolla, and Shen]{bae2023digiface}
Gwangbin Bae, Martin de La~Gorce, Tadas Baltru\v{s}aitis, Charlie Hewitt, Dong Chen, Julien Valentin, Roberto Cipolla, and Jingjing Shen.
\newblock {DigiFace-1M: 1 Million Digital Face Images for Face Recognition}.
\newblock In \emph{Proc. IEEE/CVF Winter Conference on Applications of Computer Vision}, 2023.

\bibitem[Bisogni et~al.(2022)Bisogni, Castiglione, Hossain, Narducci, and Umer]{bisogni2022impact}
Carmen Bisogni, Aniello Castiglione, Sanoar Hossain, Fabio Narducci, and Saiyed Umer.
\newblock {Impact of Deep Learning Approaches on Facial Expression Recognition in Healthcare Industries}.
\newblock \emph{IEEE Transactions on Industrial Informatics}, 18\penalty0 (8):\penalty0 5619--5627, 2022.

\bibitem[Boutros et~al.(2022)Boutros, Damer, Kirchbuchner, and Kuijper]{boutros2022elasticface}
Fadi Boutros, Naser Damer, Florian Kirchbuchner, and Arjan Kuijper.
\newblock {ElasticFace: Elastic Margin Loss for Deep Face Recognition}.
\newblock In \emph{Proc. IEEE/CVF Conference on Computer Vision and Pattern Recognition Workshops}, 2022.

\bibitem[Boutros et~al.(2023{\natexlab{a}})Boutros, Grebe, Kuijper, and Damer]{boutros2023idiff}
Fadi Boutros, Jonas~Henry Grebe, Arjan Kuijper, and Naser Damer.
\newblock {IDiff-Face: Synthetic-based Face Recognition through Fizzy Identity-Conditioned Diffusion Model}.
\newblock In \emph{Proc. IEEE/CVF International Conference on Computer Vision}, 2023{\natexlab{a}}.

\bibitem[Boutros et~al.(2023{\natexlab{b}})Boutros, Klemt, Fang, Kuijper, and Damer]{boutros2023unsupervised}
Fadi Boutros, Marcel Klemt, Meiling Fang, Arjan Kuijper, and Naser Damer.
\newblock {Unsupervised Face Recognition using Unlabeled Synthetic Data}.
\newblock In \emph{Proc. 17th International Conference on Automatic Face and Gesture Recognition}, 2023{\natexlab{b}}.

\bibitem[Crowson et~al.(2024)Crowson, Baumann, Birch, Abraham, Kaplan, and Shippole]{crowson2024scalable}
Katherine Crowson, Stefan~Andreas Baumann, Alex Birch, Tanishq~Mathew Abraham, Daniel~Z Kaplan, and Enrico Shippole.
\newblock {Scalable High-Resolution Pixel-Space Image Synthesis with Hourglass Diffusion Transformers}.
\newblock \emph{arXiv preprint arXiv:2401.11605}, 2024.

\bibitem[Crum et~al.(2023)Crum, Tinsley, Boyd, Piland, Sweet, Kelley, Bowyer, and Czajka]{crum2023explain}
Colton~R. Crum, Patrick Tinsley, Aidan Boyd, Jacob Piland, Christopher Sweet, Timothy Kelley, Kevin Bowyer, and Adam Czajka.
\newblock {Explain To Me: Salience-Based Explainability for Synthetic Face Detection Models}.
\newblock \emph{IEEE Transactions on Artificial Intelligence}, pages 1--12, 2023.

\bibitem[Daza et~al.(2023)Daza, Gomez, Morales, Fierrez, Tolosana, Cobos, and Ortega-Garcia]{daza2023matt}
Roberto Daza, Luis~F Gomez, Aythami Morales, Julian Fierrez, Ruben Tolosana, Ruth Cobos, and Javier Ortega-Garcia.
\newblock {MATT: Multimodal Attention Level Estimation for e-learning Platforms}.
\newblock In \emph{Proc. AAAI Workshop on Artificial Intelligence for Education}, 2023.

\bibitem[Deandres-Tame et~al.(2024)Deandres-Tame, Tolosana, Vera-Rodriguez, Morales, Fierrez, and Ortega-Garcia]{deandrestame2024how}
Ivan Deandres-Tame, Ruben Tolosana, Ruben Vera-Rodriguez, Aythami Morales, Julian Fierrez, and Javier Ortega-Garcia.
\newblock {How Good Is ChatGPT at Face Biometrics? A First Look Into Recognition, Soft Biometrics, and Explainability}.
\newblock \emph{IEEE Access}, 12:\penalty0 34390--34401, 2024.

\bibitem[Deng et~al.(2019)Deng, Guo, Xue, and Zafeiriou]{deng2019arcface}
Jiankang Deng, Jia Guo, Niannan Xue, and Stefanos Zafeiriou.
\newblock {ArcFace: Additive Angular Margin Loss for Deep Face Recognition}.
\newblock In \emph{Proc. IEEE/CVF Conference on Computer Vision and Pattern Recognition}, 2019.

\bibitem[Du et~al.(2022)Du, Shi, Zeng, Zhang, and Mei]{du2022elements}
Hang Du, Hailin Shi, Dan Zeng, Xiao-Ping Zhang, and Tao Mei.
\newblock {The Elements of End-to-end Deep Face Recognition: A Survey of Recent Advances}.
\newblock \emph{ACM Comput. Surv.}, 54\penalty0 (10s), 2022.

\bibitem[Duta et~al.(2021)Duta, Liu, Zhu, and Shao]{duta2021improved}
Ionut~Cosmin Duta, Li Liu, Fan Zhu, and Ling Shao.
\newblock {Improved Residual Networks for Image and Video Recognition}.
\newblock In \emph{Proc. 25th International Conference on Pattern Recognition}, 2021.

\bibitem[Erak$\iota$n et~al.(2021)Erak$\iota$n, Demir, and Ekenel]{erakιn2021recognizing}
Mustafa~Ekrem Erak$\iota$n, U{\u{g}}ur Demir, and Haz$\iota$m~Kemal Ekenel.
\newblock {On Recognizing Occluded Faces in the Wild}.
\newblock In \emph{Proc. International Conference of the Biometrics Special Interest Group}, 2021.

\bibitem[Garaev et~al.(2023)Garaev, Smirnov, Galyuk, and Lukyanets]{garaev2023facemixa}
Nikita Garaev, Evgeny Smirnov, Vasiliy Galyuk, and Evgeny Lukyanets.
\newblock {FaceMix: Transferring Local Regions for Data Augmentation in Face Recognition}.
\newblock In \emph{Proc. Neural Information Processing}, 2023.

\bibitem[Gomez et~al.(2021)Gomez, Morales, Orozco-Arroyave, Daza, and Fierrez]{gomez2021improving}
Luis~F. Gomez, Aythami Morales, Juan~R. Orozco-Arroyave, Roberto Daza, and Julian Fierrez.
\newblock {Improving Parkinson Detection Using Dynamic Features From Evoked Expressions in Video}.
\newblock In \emph{Proc. IEEE/CVF Conference on Computer Vision and Pattern Recognition Workshops}, 2021.

\bibitem[He et~al.(2016)He, Zhang, Ren, and Sun]{he2016deep}
Kaiming He, Xiangyu Zhang, Shaoqing Ren, and Jian Sun.
\newblock {Deep Residual Learning for Image Recognition}.
\newblock In \emph{Proc. IEEE/CVF Conference on Computer Vision and Pattern Recognition}, 2016.

\bibitem[Hu et~al.(2018)Hu, Shen, and Sun]{hu2018squeeze}
Jie Hu, Li Shen, and Gang Sun.
\newblock {Squeeze-and-Excitation Networks}.
\newblock In \emph{Proc. IEEE Conference on Computer Vision and Pattern Recognition}, 2018.

\bibitem[Kansy et~al.(2023)Kansy, Ra\"el, Mignone, Naruniec, Schroers, Gross, and Weber]{kansy2023controllable}
Manuel Kansy, Anton Ra\"el, Graziana Mignone, Jacek Naruniec, Christopher Schroers, Markus Gross, and Romann~M. Weber.
\newblock {Controllable Inversion of Black-Box Face Recognition Models via Diffusion}.
\newblock In \emph{Proc. IEEE/CVF International Conference on Computer Vision Workshops}, 2023.

\bibitem[Karras et~al.(2019)Karras, Laine, and Aila]{karras2019style}
Tero Karras, Samuli Laine, and Timo Aila.
\newblock {A Style-Based Generator Architecture for Generative Adversarial Networks}.
\newblock In \emph{Proc. IEEE/CVF conference on Computer Vision and Pattern Recognition}, 2019.

\bibitem[Karras et~al.(2020)Karras, Laine, Aittala, Hellsten, Lehtinen, and Aila]{karras2020analyzing}
Tero Karras, Samuli Laine, Miika Aittala, Janne Hellsten, Jaakko Lehtinen, and Timo Aila.
\newblock {Analyzing and Improving the Image Quality of StyleGAN}.
\newblock In \emph{Proc. IEEE/CVF Conference on Computer Vision and Pattern Recognition}, 2020.

\bibitem[Kim et~al.(2022)Kim, Jain, and Liu]{kim2022adaface}
Minchul Kim, Anil~K. Jain, and Xiaoming Liu.
\newblock {AdaFace: Quality Adaptive Margin for Face Recognition}.
\newblock In \emph{Proc. IEEE/CVF Conference on Computer Vision and Pattern Recognition}, 2022.

\bibitem[Kim et~al.(2023)Kim, Liu, Jain, and Liu]{kim2023dcface}
Minchul Kim, Feng Liu, Anil Jain, and Xiaoming Liu.
\newblock {DCFace: Synthetic Face Generation with Dual Condition Diffusion Model}.
\newblock In \emph{Proc. IEEE/CVF Conference on Computer Vision and Pattern Recognition}, 2023.

\bibitem[Li et~al.(2024)Li, Cao, Wang, Qi, Cheng, and Shan]{li2024photomaker}
Zhen Li, Mingdeng Cao, Xintao Wang, Zhongang Qi, Ming-Ming Cheng, and Ying Shan.
\newblock {PhotoMaker: Customizing Realistic Human Photos via Stacked ID Embedding}.
\newblock In \emph{Proc. IEEE/CVF Conference on Computer Vision and Pattern Recognition}, 2024.

\bibitem[Low et~al.(2023)Low, Chai, Park, Ann, and Cha]{low2023slackedface}
Cheng~Yaw Low, Jacky Chen~Long Chai, Jaewoo Park, Kyeongjin Ann, and Meeyoung Cha.
\newblock {SlackedFace: Learning a Slacked Margin for Low-Resolution Face Recognition}.
\newblock In \emph{Proc. 34th British Machine Vision Conference}, 2023.

\bibitem[Melzi et~al.(2023{\natexlab{a}})Melzi, Rathgeb, Tolosana, Vera-Rodriguez, Lawatsch, Domin, and Schaubert]{melzi2023gandiffface}
Pietro Melzi, Christian Rathgeb, Ruben Tolosana, Ruben Vera-Rodriguez, Dominik Lawatsch, Florian Domin, and Maxim Schaubert.
\newblock {GANDiffFace: Controllable Generation of Synthetic Datasets for Face Recognition with Realistic Variations}.
\newblock In \emph{Proc. IEEE/CVF International Conference on Computer Vision Workshops}, 2023{\natexlab{a}}.

\bibitem[Melzi et~al.(2023{\natexlab{b}})Melzi, Rathgeb, Tolosana, Vera-Rodriguez, Morales, Lawatsch, Domin, and Schaubert]{melzi2023synthetic}
Pietro Melzi, Christian Rathgeb, Ruben Tolosana, Ruben Vera-Rodriguez, Aythami Morales, Dominik Lawatsch, Florian Domin, and Maxim Schaubert.
\newblock {Synthetic Data for the Mitigation of Demographic Biases in Face Recognition}.
\newblock In \emph{Proc. IEEE International Joint Conference on Biometrics}, 2023{\natexlab{b}}.

\bibitem[Melzi et~al.(2023{\natexlab{c}})Melzi, Shahreza, Rathgeb, Tolosana, Vera-Rodriguez, Fierrez, Marcel, and Busch]{melzi2023multi}
Pietro Melzi, Hatef~Otroshi Shahreza, Christian Rathgeb, Ruben Tolosana, Ruben Vera-Rodriguez, Julian Fierrez, S\'ebastien Marcel, and Christoph Busch.
\newblock {Multi-IVE: Privacy Enhancement of Multiple Soft-Biometrics in Face Embeddings}.
\newblock In \emph{Proc. IEEE/CVF Winter Conference on Applications of Computer Vision Workshops}, 2023{\natexlab{c}}.

\bibitem[Melzi et~al.(2024{\natexlab{a}})Melzi, Tolosana, Vera-Rodriguez, Kim, Rathgeb, Liu, DeAndres-Tame, Morales, Fierrez, Ortega-Garcia, Zhao, Zhu, Yan, Zhang, Wu, Lei, Tripathi, Kothari, Zama, Deb, Biesseck, Vidal, Granada, Fickel, F\"uhr, Menotti, Unnervik, George, Ecabert, Shahreza, Rahimi, Marcel, Sarridis, Koutlis, Baltsou, Papadopoulos, Diou, Di~Domenico, Borghi, Pellegrini, Mas-Candela, S\'anchez-P\'erez, Atzori, Boutros, Damer, Fenu, and Marras]{melzi2024frcsyna}
Pietro Melzi, Ruben Tolosana, Ruben Vera-Rodriguez, Minchul Kim, Christian Rathgeb, Xiaoming Liu, Ivan DeAndres-Tame, Aythami Morales, Julian Fierrez, Javier Ortega-Garcia, Weisong Zhao, Xiangyu Zhu, Zheyu Yan, Xiao-Yu Zhang, Jinlin Wu, Zhen Lei, Suvidha Tripathi, Mahak Kothari, Md~Haider Zama, Debayan Deb, Bernardo Biesseck, Pedro Vidal, Roger Granada, Guilherme Fickel, Gustavo F\"uhr, David Menotti, Alexander Unnervik, Anjith George, Christophe Ecabert, Hatef~Otroshi Shahreza, Parsa Rahimi, S\'ebastien Marcel, Ioannis Sarridis, Christos Koutlis, Georgia Baltsou, Symeon Papadopoulos, Christos Diou, Nicol\`o Di~Domenico, Guido Borghi, Lorenzo Pellegrini, Enrique Mas-Candela, \'Angela S\'anchez-P\'erez, Andrea Atzori, Fadi Boutros, Naser Damer, Gianni Fenu, and Mirko Marras.
\newblock {FRCSyn Challenge at WACV 2024: Face Recognition Challenge in the Era of Synthetic Data}.
\newblock In \emph{Proc. of the IEEE/CVF Winter Conference on Applications of Computer Vision Workshops}, 2024{\natexlab{a}}.

\bibitem[Melzi et~al.(2024{\natexlab{b}})Melzi, Tolosana, Vera-Rodriguez, Kim, Rathgeb, Liu, DeAndres-Tame, Morales, Fierrez, Ortega-Garcia, Zhao, Zhu, Yan, Zhang, Wu, Lei, Tripathi, Kothari, Zama, Deb, Biesseck, Vidal, Granada, Fickel, Führ, Menotti, Unnervik, George, Ecabert, Shahreza, Rahimi, Marcel, Sarridis, Koutlis, Baltsou, Papadopoulos, Diou, Domenico, Borghi, Pellegrini, Mas-Candela, Ángela Sánchez-Pérez, Atzori, Boutros, Damer, Fenu, and Marras]{melzi2024frcsyn}
Pietro Melzi, Ruben Tolosana, Ruben Vera-Rodriguez, Minchul Kim, Christian Rathgeb, Xiaoming Liu, Ivan DeAndres-Tame, Aythami Morales, Julian Fierrez, Javier Ortega-Garcia, Weisong Zhao, Xiangyu Zhu, Zheyu Yan, Xiao-Yu Zhang, Jinlin Wu, Zhen Lei, Suvidha Tripathi, Mahak Kothari, Md~Haider Zama, Debayan Deb, Bernardo Biesseck, Pedro Vidal, Roger Granada, Guilherme Fickel, Gustavo Führ, David Menotti, Alexander Unnervik, Anjith George, Christophe Ecabert, Hatef~Otroshi Shahreza, Parsa Rahimi, Sébastien Marcel, Ioannis Sarridis, Christos Koutlis, Georgia Baltsou, Symeon Papadopoulos, Christos Diou, Nicolò~Di Domenico, Guido Borghi, Lorenzo Pellegrini, Enrique Mas-Candela, Ángela Sánchez-Pérez, Andrea Atzori, Fadi Boutros, Naser Damer, Gianni Fenu, and Mirko Marras.
\newblock {FRCSyn-onGoing: Benchmarking and Comprehensive Evaluation of Real and Synthetic Data to Improve Face Recognition Systems}.
\newblock \emph{Information Fusion}, 107:\penalty0 102322, 2024{\natexlab{b}}.

\bibitem[Morales et~al.(2021)Morales, Fierrez, Vera-Rodriguez, and Tolosana]{morales2021sensitivenets}
Aythami Morales, Julian Fierrez, Ruben Vera-Rodriguez, and Ruben Tolosana.
\newblock {SensitiveNets: Learning Agnostic Representations with Application to Face Images}.
\newblock \emph{IEEE Transactions on Pattern Analysis and Machine Intelligence}, 43\penalty0 (6):\penalty0 2158--2164, 2021.

\bibitem[Moschoglou et~al.(2017)Moschoglou, Papaioannou, Sagonas, Deng, Kotsia, and Zafeiriou]{moschoglou2017agedb}
Stylianos Moschoglou, Athanasios Papaioannou, Christos Sagonas, Jiankang Deng, Irene Kotsia, and Stefanos Zafeiriou.
\newblock {AgeDB: The First Manually Collected, In-The-Wild Age Database}.
\newblock In \emph{Proc. IEEE/CVF Conference on Computer Vision and Pattern Recognition Workshops}, 2017.

\bibitem[Neto et~al.(2022)Neto, Boutros, Pinto, Damer, Sequeira, Cardoso, Bengherabi, Bousnat, Boucheta, Hebbadj, Erakın, Demir, Ekenel, De~Queiroz~Vidal, and Menotti]{neto2022ocfr}
Pedro~C. Neto, Fadi Boutros, Joao~Ribeiro Pinto, Naser Damer, Ana~F. Sequeira, Jaime~S. Cardoso, Messaoud Bengherabi, Abderaouf Bousnat, Sana Boucheta, Nesrine Hebbadj, Mustafa~Ekrem Erakın, Uğur Demir, Hazım~Kemal Ekenel, Pedro~Beber De~Queiroz~Vidal, and David Menotti.
\newblock {OCFR 2022: Competition on Occluded Face Recognition from Synthetically Generated Structure-Aware Occlusions}.
\newblock In \emph{Proc. IEEE International Joint Conference on Biometrics}, 2022.

\bibitem[Neto et~al.(2023)Neto, Caldeira, Cardoso, and Sequeira]{neto2023compressed}
Pedro~C. Neto, Eduarda Caldeira, Jaime~S. Cardoso, and Ana~F. Sequeira.
\newblock {Compressed Models Decompress Race Biases: What Quantized Models Forget for Fair Face Recognition}.
\newblock In \emph{Proc. International Conference of the Biometrics Special Interest Group}, 2023.

\bibitem[Qiu et~al.(2021)Qiu, Yu, Gong, Li, Liu, and Tao]{qiu2021synface}
Haibo Qiu, Baosheng Yu, Dihong Gong, Zhifeng Li, Wei Liu, and Dacheng Tao.
\newblock {SynFace: Face Recognition with Synthetic Data}.
\newblock In \emph{Proc. IEEE/CVF International Conference on Computer Vision}, 2021.

\bibitem[Sengupta et~al.(2016)Sengupta, Chen, Castillo, Patel, Chellappa, and Jacobs]{sengupta2016frontal}
Soumyadip Sengupta, Jun-Cheng Chen, Carlos Castillo, Vishal~M Patel, Rama Chellappa, and David~W Jacobs.
\newblock {Frontal to Profile Face Verification in the Wild}.
\newblock In \emph{Proc. IEEE/CVF Winter Conference on Applications of Computer Vision}, 2016.

\bibitem[Shahreza et~al.(2023)Shahreza, George, and Marcel]{shahreza2023synthdistill}
Hatef~Otroshi Shahreza, Anjith George, and Sébastien Marcel.
\newblock {SynthDistill: Face Recognition with Knowledge Distillation from Synthetic Data}.
\newblock In \emph{Proc. IEEE International Joint Conference on Biometrics}, 2023.

\bibitem[Shahreza et~al.(2024)Shahreza, Ecabert, George, Unnervik, Marcel, Domenico, Borghi, Maltoni, Boutros, Vogel, Damer, Ángela Sánchez-Pérez, EnriqueMas-Candela, Calvo-Zaragoza, Biesseck, Vidal, Granada, Menotti, DeAndres-Tame, Cava, Concas, Melzi, Tolosana, Vera-Rodriguez, Perelli, Orrù, Marcialis, and Fierrez]{shahreza2024sdfr}
Hatef~Otroshi Shahreza, Christophe Ecabert, Anjith George, Alexander Unnervik, Sébastien Marcel, Nicolò~Di Domenico, Guido Borghi, Davide Maltoni, Fadi Boutros, Julia Vogel, Naser Damer, Ángela Sánchez-Pérez, EnriqueMas-Candela, Jorge Calvo-Zaragoza, Bernardo Biesseck, Pedro Vidal, Roger Granada, David Menotti, Ivan DeAndres-Tame, Simone Maurizio~La Cava, Sara Concas, Pietro Melzi, Ruben Tolosana, Ruben Vera-Rodriguez, Gianpaolo Perelli, Giulia Orrù, Gian~Luca Marcialis, and Julian Fierrez.
\newblock {SDFR: Synthetic Data for Face Recognition Competition}.
\newblock In \emph{Proc. 18th IEEE International Conference on Automatic Face and Gesture Recognition}, 2024.

\bibitem[Smirnov et~al.(2022)Smirnov, Garaev, Galyuk, and Lukyanets]{smirnov2022prototype}
Evgeny Smirnov, Nikita Garaev, Vasiliy Galyuk, and Evgeny Lukyanets.
\newblock {Prototype Memory for Large-Scale Face Representation Learning}.
\newblock \emph{IEEE Access}, 10:\penalty0 12031--12046, 2022.

\bibitem[Song et~al.(2021)Song, Meng, and Ermon]{song2021denoising}
Jiaming Song, Chenlin Meng, and Stefano Ermon.
\newblock {Denoising Diffusion Implicit Models}.
\newblock In \emph{Proc. International Conference on Learning Representations}, 2021.

\bibitem[Terh{\"o}rst et~al.(2021)Terh{\"o}rst, Kolf, Huber, Kirchbuchner, Damer, Moreno, Fierrez, and Kuijper]{terhoerst2021comprehensive}
Philipp Terh{\"o}rst, Jan~Niklas Kolf, Marco Huber, Florian Kirchbuchner, Naser Damer, Aythami~Morales Moreno, Julian Fierrez, and Arjan Kuijper.
\newblock {A Comprehensive Study on Face Recognition Biases Beyond Demographics}.
\newblock \emph{IEEE Transactions on Technology and Society}, 3\penalty0 (1):\penalty0 16--30, 2021.

\bibitem[Walton et~al.(2022)Walton, Hassani, Xu, Wang, and Shi]{walton2022stylenat}
Steven Walton, Ali Hassani, Xingqian Xu, Zhangyang Wang, and Humphrey Shi.
\newblock {StyleNAT: Giving Each Head a New Perspective}.
\newblock \emph{arXiv preprint arXiv:2211.05770}, 2022.

\bibitem[Wang et~al.(2018)Wang, Wang, Zhou, Ji, Gong, Zhou, Li, and Liu]{wang2018cosface}
Hao Wang, Yitong Wang, Zheng Zhou, Xing Ji, Dihong Gong, Jingchao Zhou, Zhifeng Li, and Wei Liu.
\newblock {CosFace: Large Margin Cosine Loss for Deep Face Recognition}.
\newblock In \emph{Proc. IEEE Conference on Computer Vision and Pattern Recognition}, 2018.

\bibitem[Wang et~al.(2021)Wang, Liu, Hu, Shi, and Mei]{wang2021facex}
Jun Wang, Yinglu Liu, Yibo Hu, Hailin Shi, and Tao Mei.
\newblock {FaceX-Zoo: A PyTorch Toolbox for Face Recognition}.
\newblock In \emph{Proc. 29th ACM International Conference on Multimedia}, 2021.

\bibitem[Wang and Deng(2020)]{wang2020mitigating}
Mei Wang and Weihong Deng.
\newblock {Mitigating Bias in Face Recognition Using Skewness-Aware Reinforcement Learning}.
\newblock In \emph{Proc. IEEE/CVF Conference on Computer Vision and Pattern Recognition}, 2020.

\bibitem[Wang and Deng(2021)]{wang2021deep}
Mei Wang and Weihong Deng.
\newblock {Deep Face Recognition: A Survey}.
\newblock \emph{Neurocomputing}, 429:\penalty0 215--244, 2021.

\bibitem[Yi et~al.(2014)Yi, Lei, Liao, and Li]{yi2014learning}
Dong Yi, Zhen Lei, Shengcai Liao, and Stan~Z Li.
\newblock {Learning Face Representation from Scratch}.
\newblock \emph{arXiv preprint arXiv:1411.7923}, 2014.

\bibitem[Zhang et~al.(2023)Zhang, Chen, Chai, Wu, Lagun, Beeler, and De~la Torre]{zhang2023iti}
Cheng Zhang, Xuanbai Chen, Siqi Chai, Chen~Henry Wu, Dmitry Lagun, Thabo Beeler, and Fernando De~la Torre.
\newblock {ITI-GEN: Inclusive Text-to-Image Generation}.
\newblock In \emph{Proc. IEEE/CVF International Conference on Computer Vision}, 2023.

\bibitem[Zhang et~al.(2016)Zhang, Zhang, Li, and Qiao]{zhang2016joint}
Kaipeng Zhang, Zhanpeng Zhang, Zhifeng Li, and Yu Qiao.
\newblock {Joint Face Detection and Alignment Using Multitask Cascaded Convolutional Networks}.
\newblock \emph{IEEE Signal Processing Letters}, 23\penalty0 (10):\penalty0 1499--1503, 2016.

\bibitem[Zhong et~al.(2021)Zhong, Deng, Hu, Zhao, Li, and Wen]{zhong2021sface}
Yaoyao Zhong, Weihong Deng, Jiani Hu, Dongyue Zhao, Xian Li, and Dongchao Wen.
\newblock {SFace: Sigmoid-Constrained Hypersphere Loss for Robust Face Recognition}.
\newblock \emph{IEEE Transactions on Image Processing}, 30:\penalty0 2587--2598, 2021.

\bibitem[Zhou et~al.(2023)Zhou, Jia, Li, Shen, and Duan]{zhou2023uniface}
Jiancan Zhou, Xi Jia, Qiufu Li, Linlin Shen, and Jinming Duan.
\newblock {UniFace: Unified Cross-Entropy Loss for Deep Face Recognition}.
\newblock In \emph{Proc. IEEE/CVF International Conference on Computer Vision}, 2023.

\bibitem[Zhu et~al.(2021)Zhu, Huang, Deng, Ye, Huang, Chen, Zhu, Yang, Lu, Du, and Zhou]{zhu2021webface260m}
Zheng Zhu, Guan Huang, Jiankang Deng, Yun Ye, Junjie Huang, Xinze Chen, Jiagang Zhu, Tian Yang, Jiwen Lu, Dalong Du, and Jie Zhou.
\newblock {WebFace260M: A Benchmark Unveiling the Power of Million-Scale Deep Face Recognition}.
\newblock In \emph{Proc. IEEE/CVF Conference on Computer Vision and Pattern Recognition}, 2021.

\end{thebibliography}
